\pdfoutput=1
\documentclass[11pt]{article}
\usepackage{acl}
\usepackage{times}
\usepackage{latexsym}
\usepackage[T1]{fontenc}
\usepackage[utf8]{inputenc}
\usepackage{microtype}

\usepackage{amsmath}
\usepackage{amsfonts}
\usepackage{amssymb}
\usepackage{natbib}
\usepackage{microtype}
\usepackage{bbm}
\usepackage{graphicx}
\usepackage{xcolor}
\usepackage{mathtools}
\usepackage{csquotes}
\usepackage{mdframed}
\usepackage{comment}
\usepackage{enumitem}
\usepackage{amsthm}
\usepackage{booktabs}
\usepackage{rotating}
\usepackage{makecell}
\usepackage{soul}
\usepackage{fancyvrb}
\usepackage{makecell}

\theoremstyle{definition}
\newtheorem{definition}{Definition}[section]

\theoremstyle{definition}

\newenvironment{talign}
 {\align}
 {\endalign}
\newenvironment{talign*}
 {\csname align*\endcsname}
 {\endalign}

\newcommand{\modelname}{\textsc{ExSum}}

\title{\modelname{}: From Local Explanations to Model Understanding}

\author{Yilun Zhou\\ MIT CSAIL \And Marco Tulio Ribeiro \\Microsoft Research\\\texttt{\{yilun,julie\_a\_shah\}@csail.mit.edu\space\space marcotcr@microsoft.com}\\\\\url{https://yilunzhou.github.io/exsum/} \And Julie Shah\\MIT CSAIL}

\usepackage{xcolor}
\usepackage{tikz}
\usepackage{arydshln}
\usepackage{multirow}

\usepackage{pifont}
\newcommand{\cmark}{\ding{51}}
\newcommand{\xmark}{\ding{55}}

\definecolor{borderblue}{HTML}{5288E8}
\definecolor{borderpink}{HTML}{D19BB2}
\definecolor{borderc1}{HTML}{6aa84f}
\definecolor{borderc2}{HTML}{f1c232}
\definecolor{borderc3}{HTML}{e69138}

\newcommand{\ctext}[3][RGB]{%
  \begingroup
  \definecolor{hlcolor}{#1}{#2}\sethlcolor{hlcolor}%
  \hl{#3}%
  \endgroup
}

\newcommand{\ffbox}[2]{%
  {% open a group for a local setting
   \setlength{\fboxsep}{-1.35\fboxrule}% the rule will be inside the box boundary
   \fcolorbox{#1}{white}{\hspace{1.2pt}\raisebox{0.5pt}\strut#2\hspace{1.2pt}}% print the box, with some padding at the left and right
  }% close the group
}

\newcommand{\x}{\phantom{0}}

\begin{document}
\maketitle

\setlength{\abovedisplayskip}{1ex}
\setlength{\belowdisplayskip}{1ex}
\setlength{\abovedisplayshortskip}{1ex}
\setlength{\belowdisplayshortskip}{1ex}

\begin{abstract}

Interpretability methods are developed to understand the working mechanisms of black-box models, which is crucial to their responsible deployment. Fulfilling this goal requires both that the explanations generated by these methods are correct \textit{and} that people can easily and reliably understand them. While the former has been addressed in prior work, the latter is often overlooked, resulting in informal model understanding derived from a handful of local explanations. In this paper, we introduce explanation summary (\modelname{}), a mathematical framework for quantifying model understanding, and propose metrics for its quality assessment. On two domains, \modelname{} highlights various limitations in the current practice, helps develop accurate model understanding, and reveals easily overlooked properties of the model. We also connect understandability to other properties of explanations such as human alignment, robustness, and counterfactual minimality and plausibility. 

\end{abstract}

\section{Introduction}
\label{sec:intro}

Understanding a model's behavior is often a prerequisite for deploying it in the real world, especially in high-stake scenarios such as financial, legal, and medical domains. Unfortunately, most high-performing models, such as neural networks, are black-boxes. Thus, model-agnostic interpretability techniques have been developed, with the majority being ``local'' -- algorithms that produce an explanation for a specific input at a time \citep[e.g.,][]{li2015visualizing, ribeiro2016should}. 

Even with these local explanations, there are still two hurdles to overcome before achieving the ultimate goal of complete understanding of a model. First, some local explanations may not correctly (or faithfully) represent the model's reasoning process \citep{jacovi2020towards}, as has been demonstrated both theoretically \citep{nie2018theoretical} and empirically \citep{adebayo2018sanity} in prior work. As a result, correctness evaluation has received much attention in the community \citep[e.g.,][]{samek2016evaluating, arras2019evaluating, zhou2021feature}. 

\begin{figure}[!t]
    \centering
    \includegraphics[width=\columnwidth]{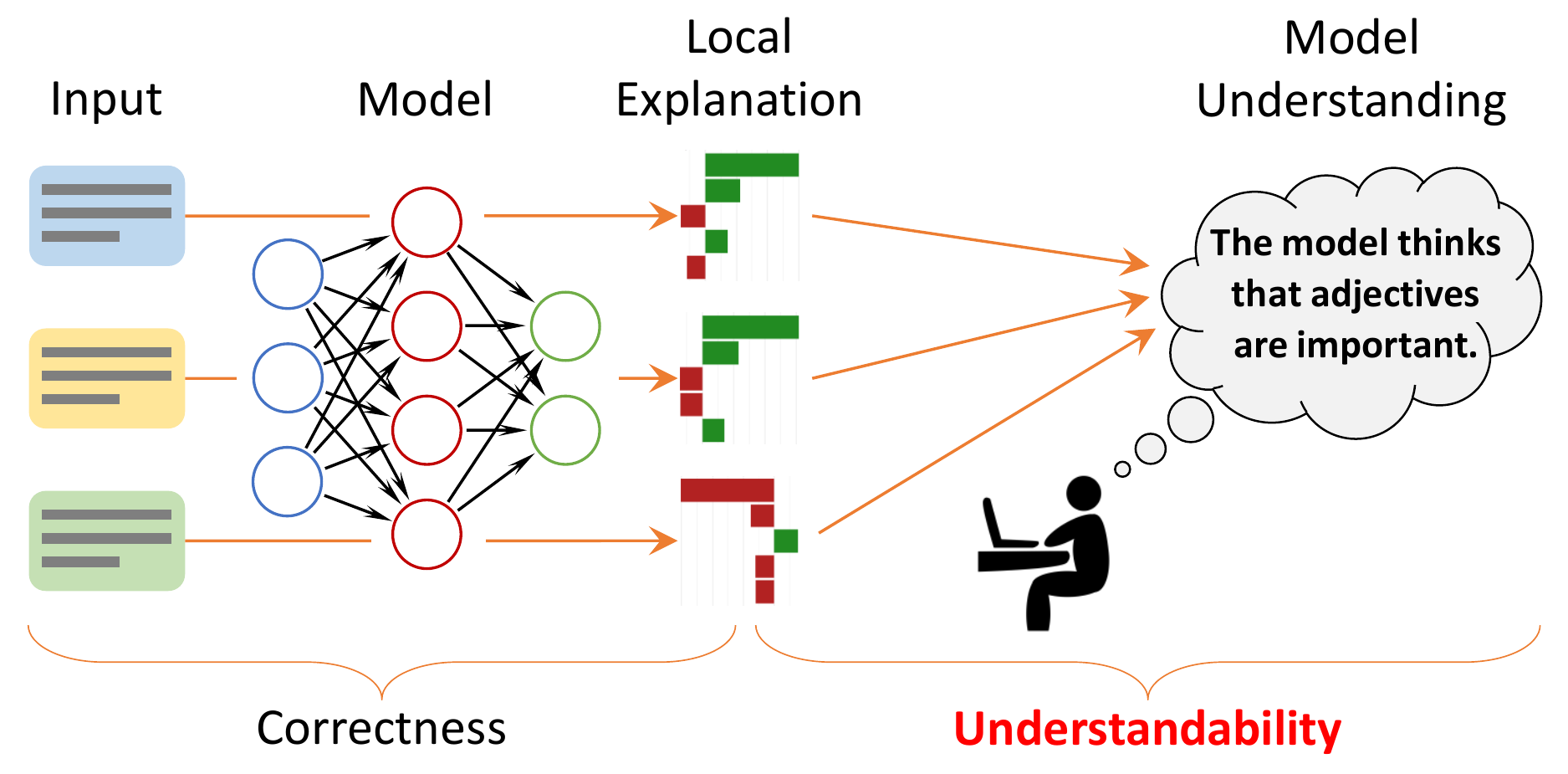}
    \caption{Local model explanations need to be both correct and easily understandable. While much prior work \citep[e.g.,][]{zhou2021feature} has studied the former property, this paper focuses on the latter, which has thus far been largely ignored. }
    \label{fig:0}
\end{figure}

Another mostly overlooked property of explanations is their \textit{understandability}. As the model understanding pipeline depicted in Fig.~\ref{fig:0} shows, explanations need to be both correct and easily understandable, since even correct explanations are not as valuable if they lead to incorrect understanding. However, the concept of understandability has yet to be formalized, and instead users often derive model understanding from few examples in a non-rigorous (and potentially incorrect) manner. 

\begin{figure*}[htbp]
    \centering
    \includegraphics[width=\textwidth]{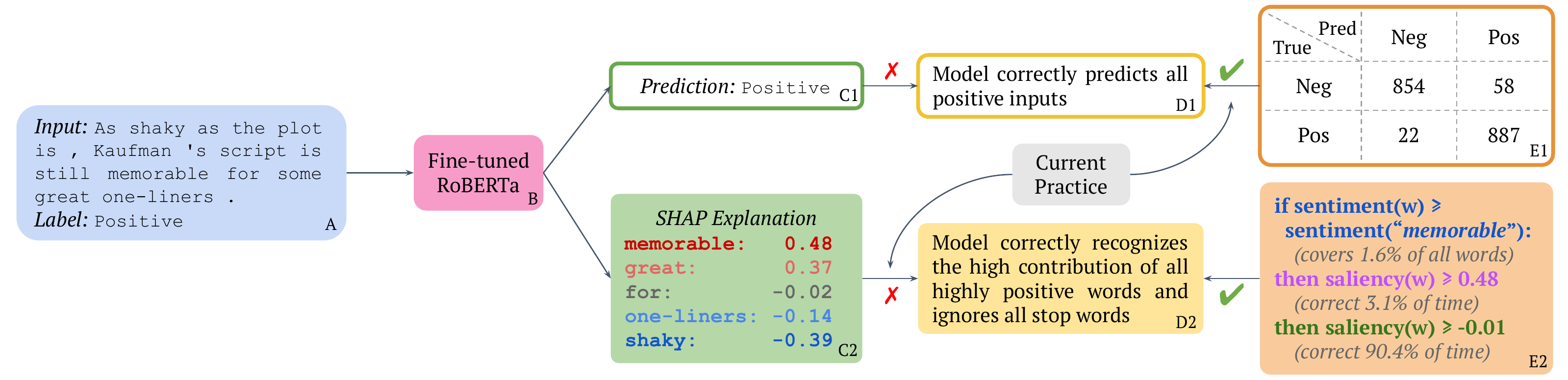}
\caption{An analogy between understanding model prediction (top route) and model explanation (bottom route). A test input {\footnotesize \ctext[RGB]{201,218,248}{(A)}} is fed into a fine-tuned RoBERTa model {\footnotesize \ctext[RGB]{247,156,201}{(B)}}, which generates a correct prediction {\footnotesize \ffbox{borderc1}{(C1)}} and reasonable explanation {\footnotesize \ctext[RGB]{182,215,168}{(C2)}}. While generalized claims of understanding model \textit{performance} {\footnotesize \ffbox{borderc2}{(D1)}} are made rigorously from quantitative statistics such as the test set confusion matrix {\footnotesize \ffbox{borderc3}{(E1)}}, claims of understanding model \textit{behavior} \ctext[RGB]{255,230,153}{(D2)} are predominantly derived informally from one or few explanations {\footnotesize \ctext[RGB]{182,215,168}{(C2)}}. In this paper, we argue the necessity of formalizing this process, and propose the explanation summary (\modelname{}) framework {\footnotesize \ctext[RGB]{249,203,156}{(E2)}}, which reveals the severe limitations of the \textit{ad hoc} model understanding {\footnotesize \ctext[RGB]{255,230,153}{(D2)}}. }
\label{fig:1}
\end{figure*}

Consider the sentiment classification task shown in Fig.~\ref{fig:1}. On a test input, the model makes the correct prediction of positive sentiment. Obviously, this evidence is insufficient to conclude that ``\textit{in general}, the model classifies positive inputs correctly'', because even a random-guess model is correct 50\% of the time on a single instance. Instead, statistics such as the confusion matrix serve to rigorously support (or refute) generalization claims about model \textit{performance} -- for example, ``the model is correct 97.6\% of the time on positive inputs'' -- ensuring an accurate understanding of model performance.

Do we understand model \textit{behaviors} in the same rigorous way? Fig.~\ref{fig:1} shows that the SHAP score \cite{lundberg2017unified} of the word ``\textit{memorable}'' is highest at 0.48, while that of ``\textit{for}'' is negligible at -0.02. Therefore, it is tempting to conclude that ``\textit{in general}, the model recognizes the high positive contribution of highly positive words and ignores stop words'' -- as expected for an accurate sentiment classifier. However, this is a generalization from a single instance, and thus potentially unreliable. We need the ``confusion matrix'' analogue for such claims, which to the best of our knowledge does not exist, making it hard to derive model understanding from local explanations.

In this paper, we propose \modelname{}, a mathematical framework to formalize model understanding. In \modelname{}, each piece of ``model understanding'' is specified precisely via a rule that links inputs to attribution values. For example, the tentative understanding described in the previous paragraph could be formalized as ``words more positive than \textit{memorable} (as measured by the word sentiment score given in the dataset, e.g., \textit{flawless}, \textit{charming}, etc) have SHAP attribution value in the [0.48, 1] range.'' This precise definition allows for quantitative evaluations. For example, this rule covers 1.6\% of all words in the corpus, and is only correct 3.1\% of the time. For the rule to be 90\% correct, we need a wide and uninformative range of [-0.01, 1], indicating that a hasty generalization from ``\textit{memorable}'' is unwarranted.
Similarly, a saliency range of [-0.05, 0.05] for stop words is only correct 64\% of the time: over 1/3 of stop words have \textit{non}-negligible saliency -- an understanding that is easily available with \modelname{}, but might be missed with informal explanation inspection.
We define metrics to establish the quality profile of each rule and present a tool that makes it easy for users to construct \modelname{} rules from local explanations. Finally, we demonstrate how \modelname{} reveals the various drawbacks in the current practices of \textit{ad hoc} model understanding, and allows for better understanding of model behavior in two separate tasks.

\section{On Generalized Model Understanding}
\label{sec:eoe-framework}

Besides the practical example above, we start from first principles and argue that \textit{generalized} model understanding is the central concept for explanation usefulness. Local explanations are \textit{\ul{mathematical descriptions} (MD) of some aspect of model behavior, for specific inputs}. For example, gradient saliency (in the embedding space) is the sensitivity of the prediction to infinitesimal changes in the token embedding; occlusion saliency is the prediction change if individual embeddings are zeroed out. It is with these mathematical descriptions that people associate \textit{\ul{high-level interpretations} (HL) of model behavior}, such as associating the above two metrics with word importance. This (unconscious) train of thought can be described as follows:
\begin{align*}
x \rightarrow \mathrm{MD} \rightarrow \mathrm{HL}.
\end{align*}

Crucially, people rarely study MD or HL for one \emph{specific} input, as explanations are often used to understand broader model behaviors, such as reliance upon spurious correlation, non-discrimination of a protected class, or usage of unknown scientific principles. We elaborate upon these use cases in App.~\ref{app:real-world} to demonstrate that people implicitly or explicitly seek generalized model understanding. 
From another perspective, analogous to why people ultimately focus on the \textit{generalization} accuracy of a model, they (should) focus on \textit{generalized} model understanding derived from local explanations. 

For example, after observing that \textit{some} highly polar words have high contribution for a sentiment classification model, people conclude that \textit{all} highly polar words have high contribution. This process can be formalized as follows: 
\begin{align*}
    \left. \begin{tabular}{c}
         $x_1 \rightarrow \mathrm{MD}_1 \rightarrow \mathrm{HL}_1$ \\
         ... \\
         $x_n \rightarrow \mathrm{MD}_n \rightarrow \mathrm{HL}_n$ 
    \end{tabular}
    \right\}\rightarrow \mathrm{HL}^{(\mathrm{g})}, 
\end{align*}
\noindent where $\mathrm{HL}^{(\mathrm{g})}$ is the \textit{generalized} high-level model understanding. This generalization is too informal, not least because the step from $\mathrm{MD}_i$ to $\mathrm{HL}_i$ is itself already informal. Alternatively, we propose to generalize at the MD level, as follows: 
\begin{align*}
    \left. \begin{tabular}{c}
         $x_1 \rightarrow \mathrm{MD}_1$ \\
         ... \\
         $x_n \rightarrow \mathrm{MD}_n$ 
    \end{tabular}
    \right\}\rightarrow \mathrm{MD}^{(\mathrm{g})} \rightarrow \mathrm{HL}^{(\mathrm{g})}. 
\end{align*}
\noindent Since MDs are rigorously defined mathematical quantities (e.g., the prediction of the sentence drops by 32\% after the embedding of ``great'' is zeroed out), we can define and evaluate the quality of their generalization, and $\mathrm{HL}^{(\mathrm{g})}$ can also include any failures and anomalies. 
As each MD is a local explanation, we call $\mathrm{MD}^{(\mathrm{g})}$ the \textit{explanation summary} (\modelname{}), and proceed by instantiating this principle for feature attribution explanations.

\section{The \modelname{} Framework}
\label{sec:eoe-rules}

\subsection{Setup and Notation}

We focus on the classification setting, but all the ideas below can extend straightforwardly to regression. We have an input space $\mathcal X$ and output space $\mathcal Y=\{1, ..., K\}$ of $K$ classes. A data point is an input-output pair $d=(x, y)\in \mathcal D=\mathcal X\times \mathcal Y$, distributed as $\mathbb P_D$. We consider a model $m: \mathcal X\rightarrow \Delta^{K-1}$ where $m(x)$ is the predicted class distribution on the probability simplex. 

Feature attribution explainers assign an attribution, also known as saliency or importance, to each input feature, such as a token in a text input. For an instance $(x, y)$, each feature of $x$ is called a fundamental explanation unit (FEU), defined as $u=(x, y, l)\in \mathcal U$ with $1\leq l\leq L_x$ as the feature index. $e(u)\in \mathcal E$ represents the attribution value assigned to it, where $\mathcal E$ is the attribution space, such as $[-1, 1]$ for normalized explanations. $e(u_-)=\left(e_x^{(1)}, ..., e_x^{(l-1)}, e_x^{(l+1)}, ..., e_{x}^{(L_x)}\right)\in \mathcal E^*_-$ denotes the explanations on all other FEUs of $x$. 

We \textit{define} a distribution $\mathbb P_U$ over $\mathcal U$ such that the probability (or probability density) of $u=(x, y, l)$ is $1/L_x$ of that of $d$ under the data distribution $\mathbb P_D$. In other words, sampling of $u$ can be performed in two steps: first draw an instance $d=(x, y)\sim \mathbb P_D$, then a feature index $l\sim \mathrm{Unif}(\{1, ..., L_x\})$.

\subsection{\modelname{} Rules}

An \modelname{} rule formalizes a piece of model understanding, such as that for positive words in Fig.~\ref{fig:1}, which we use as the running example. 

\begin{definition}[\modelname{} rule]
An \modelname{} rule $r$ is defined by two functions. A binary-valued \textit{applicability function} $a: \mathcal U\rightarrow \{0, 1\}$ determines whether the rule applies to a given FEU, with 1 being applicable and 0 otherwise. We use $a(\mathcal U) = \{u\in \mathcal U: a(u)=1\}$ to denote the \textit{applicability set}. A set-valued \textit{behavior function} is defined as $b: a(\mathcal U) \times \mathcal E^*_{-} \rightarrow \mathcal P(\mathcal E)$ where $\mathcal P(\mathcal E)$ is the power set (i.e., the set of all subsets) of $\mathcal E$. This function predicts a set of possible explanation values for the FEU, called the \textit{behavior range}. The rule is written as $r=\langle a, b\rangle$. We abbreviate $b(u, e(u_-))$ as $b(u)$ and refer to the two functions as $a$- and $b$-functions. 
\end{definition}

For FEU $u=(x, y, l)$, the $a$-function typically depends only on $x_l$, but could depend on the entire input $x$ (e.g., for long sentences) or the output $y$ (e.g., for positive class). In our example, it tests whether the sentiment score is greater than that of the word ``\textit{memorable}'' (0.638). The $b$-function usually outputs a constant range. Since ``\textit{memorable}'' has a saliency of 0.479, the range is [0.479, 1.0]. 

\subsection{Additional Examples}
While we expect most rules to use rather simple $a$- and $b$-functions, they can also be more complex with more nuanced aspects. For the following examples, recall that $u=(x, y, l)$. An applicability function can target words only in long sentences using a conjunction with $\mathrm{len}(x)\geq L$, where $L$ is the threshold. We can also target inputs with ambivalent predictions with $\max_c m(x)_c\leq 0.6$, where $\max_c m(x)_c$ is the probability of the predicted class. For behavior functions, to indicate the first word of the sentence has higher saliency than the rest, we can define $b(u, e_-)=(\max_{l'\geq 2}e_{-}^{(l')}, 1.0]$, where the $a$-function selects the first word (i.e. $a(u)=\mathbbm{1}_{l=1}$). Similarly, to describe that an FEU has higher saliency than all the verbs in a sentence, we can can use $b(u, e_-) = \left(\max_{\,l': \mathrm{is\_verb}(x_{l'})} \{e_-^{(l')}\}, \allowbreak +\infty\right)$.

\subsection{\modelname{} Rule Unions}
Since a single \modelname{} rule is designed to capture one aspect of model understanding, multiple rules are often necessary for comprehensive understanding. However, conflicts can occur when multiple rules apply to the same FEU but the $b$-functions are different. We resolve them by defining the composition of two or more rules into a \textit{rule union}. 

\begin{definition}[Precedence-Mode Composition]
Two rules, $r=\langle a, b\rangle$ and $r'=\langle a', b'\rangle$, can be composed into a precedence-mode rule union $r^* = r > r'$ defined as $r^* = \langle a^*, b^*\rangle$ where
\begin{align}
    a^*(u) &= \mathbbm 1\{a(u) + a'(u)\geq 1\}, \\
    b^*(u) &= \begin{cases}
    b(u) \quad &\mathrm{if} \,\, a(u)=1, \\
    b'(u) \quad &\mathrm{if} \,\, a(u)=0, a'(u)=1, 
    \end{cases}
\end{align}
represent the $a$- and $b$-functions of rule union $r^*$, with semantics similar to those for rules. 
\end{definition}

For example, if we want to split positive adjectives into a separate rule from other positive words, we create a rule to test for part-of-speech and sentiment score, and assign a higher precedence to this rule, such that the original rule is only applicable to the remaining non-adjectives. One useful practice is to include a lowest-precedence catch-all rule that covers everything not addressed by other rules, with a constant $a(u)=1$ function, which leaves no FEUs unaccounted for. 

\begin{definition}[Intersection-Mode Composition]
Two rules, $r=\langle a, b\rangle$ and $r'=\langle a', b'\rangle$, can be composed into an intersection-mode rule union $r^* = r \mathbin{\&} r'$ defined as $r^* = \langle a^*, b^*\rangle$ where
\begin{align}
    a^*(u) &= \mathbbm 1\{a(u) + a'(u)\geq 1\}; \\
    b^*(u) &= \begin{cases}
    b(u) \quad &\mathrm{if} \,\, a(u)=1, a'(u)=0, \\
    b'(u) \quad &\mathrm{if} \,\, a(u)=0, a'(u)=1, \\
    b(u) \cap b'(u) \quad &\mathrm{if} \,\, a(u)=a'(u)=1. 
    \end{cases}
\end{align}
\end{definition}

Unlike precedence-mode, intersection-mode composition is symmetric with respect to the two rules. This mode is helpful when each property of an FEU has a corresponding behavior range, and the final behavior range of an FEU depends on FEU's properties. For example, if verbs have a behavior range of [-0.4, 0.4] and strongly positive words have a behavior range of [0.3, 1], a strongly positive verb would have a behavior range [0.3, 0.4], or the intersection of the two constituent ranges. In our case studies, however, we do not encounter any situations in which intersection-mode compositions were preferable. 

Since rule unions are also defined by $a$- and $b$-functions, they can form other rule unions in the same way. Recursively, this results in a list of rules composed into a single rule union, written as $r^*=(r_3 > r_1) \mathbin{\&} ((r_4 \mathbin{\&} r_2) > r_5)$. This rule union represents our \textit{generalized model understanding}. 

\subsection{Quality Metrics}

We propose three metrics for establishing the quality profiles of \modelname{} rules or rule unions. 

\begin{definition}[Coverage]
The coverage of a rule (union) $r=\langle a, b\rangle$ is defined as follows:
\begin{talign}
\kappa(r) = \mathbb E_{U\sim \mathbb P(U)}\left[a(U)\right]. 
\end{talign}
\end{definition}

This represents the fraction of FEUs that we attempt to understand. While individual rules may have low coverage because they specialize in aspects of the model behavior, we want their union to have high coverage to achieve a comprehensive understanding of the model and prevent model prediction from being excessively affected by the uncovered (i.e. unexplained) input features. For our positive word rule, the coverage is the frequency of those words in the corpus and not surprisingly is only 1.6\%. By contrast, including a catch-all rule in the union maxes out its coverage value at 100\%. 

\begin{definition}[Validity]
Let $\mathbb P_{a(U)}$ be $\mathbb P_U$ truncated to the set of applicable FEUs. The validity of a rule (union) $r=\langle a, b\rangle$ is then defined as follows, capturing the intuitive notion of a ``correct'' understanding:
\begin{talign}
\nu(r) = \mathbb E_{U\sim \mathbb P_{a(U)}}\left[\mathbbm 1\{e(U)\in b(U)\}\right]. 
\end{talign}
\end{definition}

For our example, we compute it as the frequency that the saliency of those words is actually in the range of [0.479, 1] -- which turns out to be only 3.1\% of the time. However, validity alone is not sufficient, as it increases with wider behavior range. We thus establish sharpness as a competing metric. 

\begin{definition}[Sharpness]
Let $\mathbb P_E$ be the probability measure corresponding to the marginal distribution over explanation values generated by the explainer on $u\sim \mathbb P_U$. The sharpness of a rule (union) $r=\langle a, b\rangle$ is defined as follows:
\begin{talign}
\sigma(r) = \mathbb E_{U\sim P_{a(U)}}\left[1 - \mathbb P_E(b(U)_{\backslash U})\right], 
\end{talign}
where $b(U)_{\backslash U} = b(U) \backslash \{U\}$ removes the actual attribution value $U$ from the behavior range to prevent penalizing sharpness simply because the attribution value is very common (e.g., zero for sparse explanations), in which case $\mathbb P_E$ is discrete at $U$. 
\end{definition}
Sharpness represents precision in the understanding, as $1{-}\sigma(r)$ gives the probability that a random FEU explanation value is correct. Thus, a lack of precision represented by a wide behavior range has minimal sharpness. We use the probability measure $\mathbb P_{E}$ to define the ``size'', as it is consistent across all explanation distributions, most of which are non-uniform. A more general interpretation of sharpness is the consistency of the described model behavior: if a behavior range is wide (e.g., containing very positive \textit{and} negative saliencies), then it is less sharp, and hence less useful. $\mathbb P_E$ could be replaced by an application-specific diversity measure, though the precision notion may be lost. 

There is generally a trade-off between validity and sharpness, as more precise rules (i.e., those with narrower behavior ranges) are less likely to be valid. For our rule, the probability of a \textit{random} word saliency being in [0.479, 1.0] is 0.2\%, indicating that explanation values are rarely higher than 0.479. This makes sharpness very high at 99.8\%. However, the rule is not useful because of its low validity; i.e., it is almost never correct. By comparison, the looser range of [-0.01, 1.0] has 90.4\% validity but 28.6\% sharpness. There is another trade-off between coverage and the two, since a larger set of covered FEUs tends to be more diverse, making it harder to write a $b$-function that remains as valid and sharp simultaneously. 

Since these metrics are all expected values, we can estimate them by their empirical estimate from a dataset (i.e., a simple average), and $\mathbb P_E$ can be constructed by kernel density estimation.

\section{\modelname{} Development Process and GUI}
\label{sec:gui}

\begin{figure}[!b]
    \centering
    \includegraphics[width=\columnwidth]{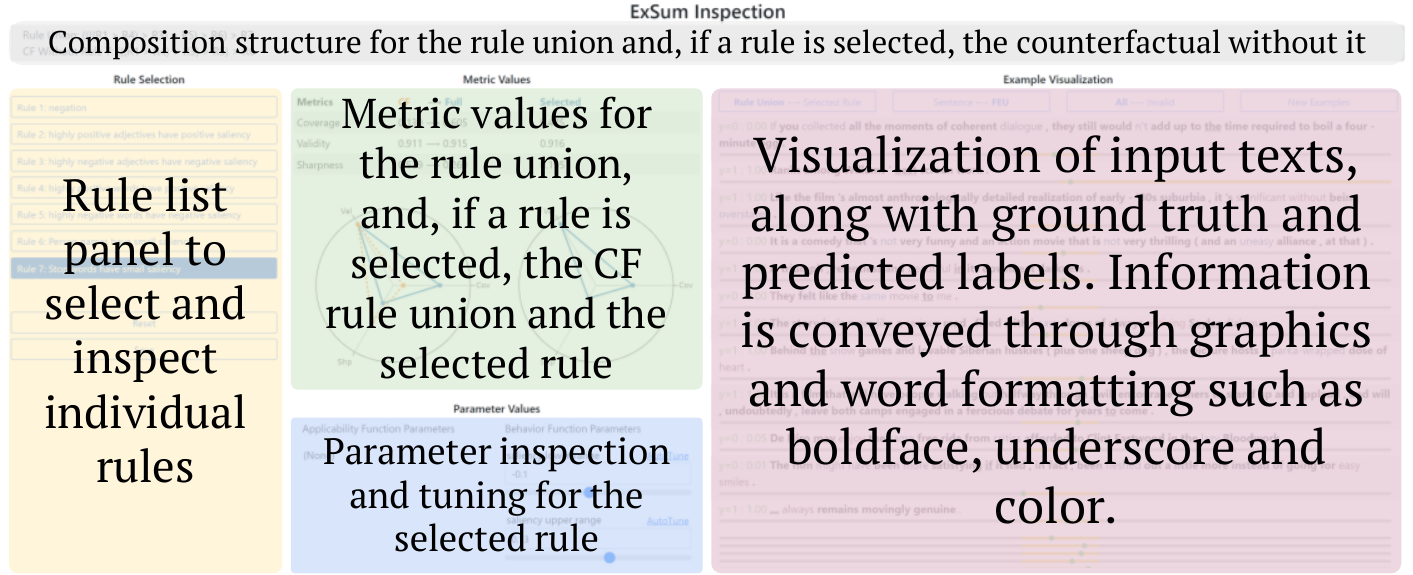}
    \caption{\modelname{} inspection GUI. }
    \label{fig:gui-overview}
\end{figure}

We describe a systematic procedure for authoring \modelname{} rule unions from scratch and utilize it in Sec.~\ref{sec:experiment}. Starting from an empty rule union with no FEUs covered, we iteratively create rules that target uncovered FEUs. Each rule describes one model behavior, such as that for highly positive words. For a rule, the $a$- and $b$-functions need to be defined, which may involve setting and tuning parameters, such as the sentiment threshold. Last, we add a lowest precedence catch-all rule if any FEUs remain uncovered. During this process, we may also merge or split rules and change the composition structure according to the metric values. 

To support these steps, we developed a Python Flask-based~\citep{grinberg2018flask} graphical user interface (GUI, Fig.~\ref{fig:gui-overview}). 
Users can visualize the FEUs, with font formatting for their coverage and validity. Users can also filter for uncovered or invalid FEUs, iteratively constructing and refining the rule union.
\modelname{} rule definitions usually include parameters such as the sentiment threshold. Manually selecting correct values for the parameters is tedious, so the lower middle panel of the GUI implements automatic parameter tuning for a given target metric value. Installation and usage instructions for the GUI are available on the project page\footnote{\url{https://yilunzhou.github.io/exsum/}}.

% \vspace{-0.5ex}
\section{Evaluation}
% \vspace{-0.5ex}
\label{sec:experiment}
We construct \modelname{} rule unions for SST and QQP models (details in App.~\ref{app:experiment}). 
We split the test set into a \textit{construction set} to create the rule union and tune its parameters (analogous to the training and validation set in supervised model training) and an \textit{evaluation set} to compute unbiased estimates of the metric values (analogous to the test set).

% \vspace{-0.5ex}
\subsection{Sentiment Classification}
% \vspace{-0.5ex}
\label{sec:experiment-1}
\paragraph{Setup}
We use SHAP explanations \citep{lundberg2017unified} for fine-tuned RoBERTa \citep{liu2019roberta}, and take 300 random sentences as the construction set, with the remaining 1910 sentences as the evaluation set. We compute five features for each FEU: sentiment score, part of speech (POS), named entity recognition (NER), dependency tag (DEP) and word frequency. For example, the word ``same'' in the sentence ``\textit{They felt like the \ul{same} movie to me .}'' has sentiment score of 0.028, POS$\,=\,$ADJ, NER$\,=\,$O, DEP$\,=\,$amod, and frequency of 7.14e-4, with SHAP saliency of -0.82. 

\paragraph{Current Practice}

We evaluate the current practice of extracting \textit{informal} model understanding from local explanation inspection against the three metrics. We assess three values of $K$, the number of inspected instances: 1, for the typical \textit{ad hoc} setting of generalization from a single explanation, 10, for a more careful investigation, and 30, which is quite cumbersome for manual inspection. These examples are selected either randomly or by submodular pick \citep{ribeiro2016should}. Next, we consider three ways to extract model understanding -- belief-guided (BG), quantile-fitting (QF) and word-level (WL) -- and apply them to create rules on strongly positive words and stop words introduced in Sec.~\ref{sec:intro}. For the strongly positive word rule, BG mandates that words more positive than the average sentiment score should have an above-average saliency score, representing the belief of a positive correlation between the two. For the stop word rule, a saliency range belief of [-0.05, 0.05] is averaged with the observed range. For both rules, QF extracts the 5\%-95\% quantile interval of the saliencies for words covered by the respective rule. WL, by contrast, creates a behavior range \emph{for each word seen}, with 0.03 margin on both sides. App.~\ref{app:bg-qf-wl} presents technical details for these.

\begin{figure}[!t]
    \centering
    \includegraphics[width=0.8\columnwidth]{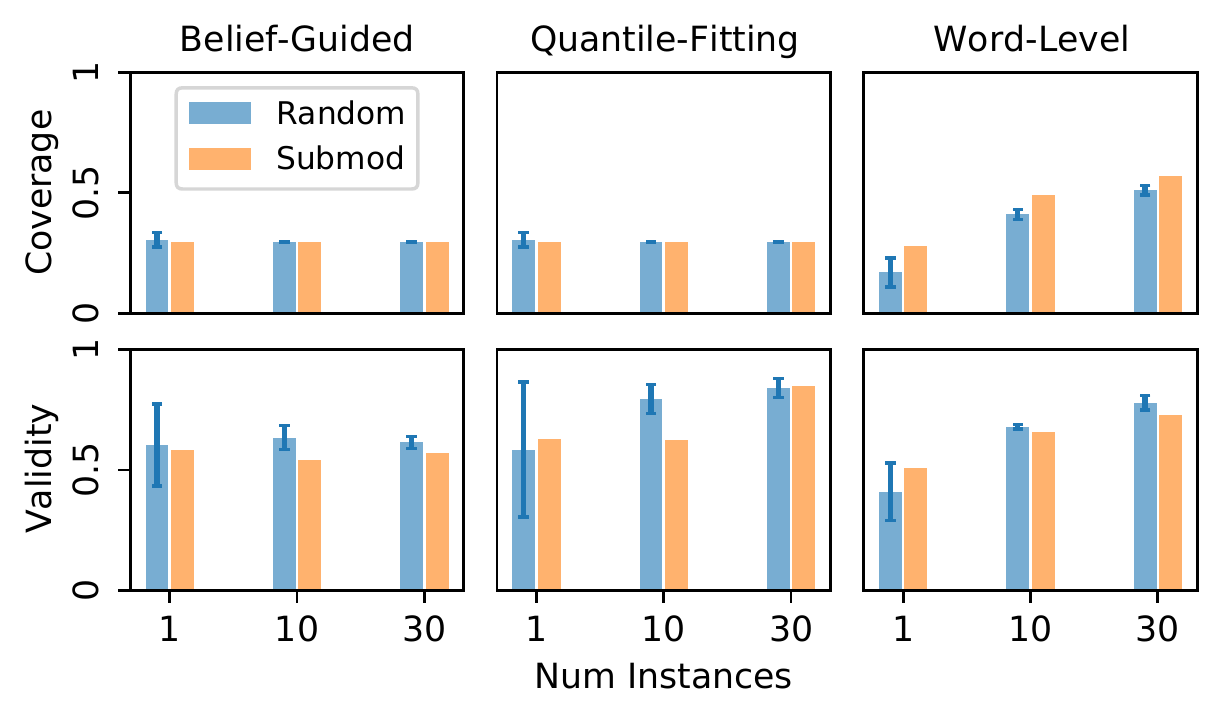}
    
    \caption{Coverage and validity metrics for the three current practice modes. Tab.~\ref{tab:current-practice-sst} of App.~\ref{app:bg-qf-wl} presents the complete numerical data (also with sharpness). }
    
    \label{fig:current-practice-sst}
\end{figure}

We formalize the understanding derived from the selected instances and plot their coverage and validity metrics on the evaluation set in Fig.~\ref{fig:current-practice-sst}. For BG and QF, the bars represent the average metric value of the positive word and stop word rules. For WL, the bars represent the metric for the rule union consisting of an individual rule for each unique word. Error bars for the random pick represent the standard deviation across five iterations. Tab.~\ref{tab:current-practice-sst} of App.~\ref{app:bg-qf-wl} presents the complete statistics for all metric values, and we highlight several findings. 

\begin{figure}[!t]
    \centering
    
    \includegraphics[width=0.8\columnwidth]{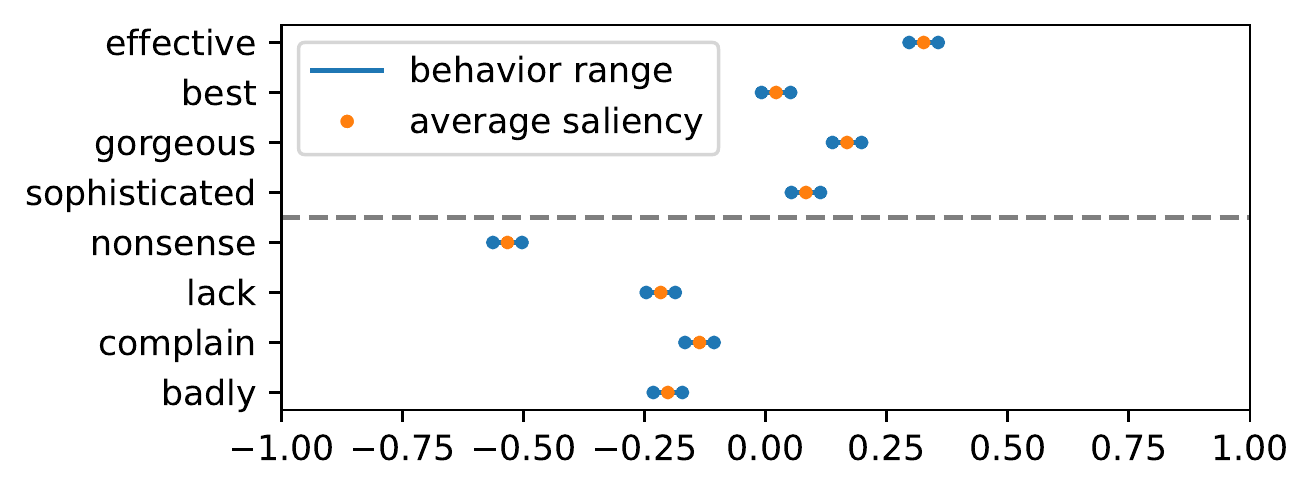}
    
    \caption{Behavior ranges can vary widely and unpredictably on similar words for WL rules. }
    
    \label{fig:sst-baseline}
\end{figure}

\begin{itemize}[leftmargin=*, parsep=0pt, itemsep=0pt, topsep=0pt, partopsep=0pt]
    \item A very small number of samples (e.g., 1) exhibit large variance for random pick, and low validity for both pick methods. This confirms the intuition that model understanding from very few explanations should be avoided. 
    \item BG overall yields low validity, because its ``beliefs'' turn out to be quite incorrect. This suggests a strong prior belief about how the model works could lead to incorrect conclusions. 
    \item While submodular pick can select a more diverse set of words, to the particular benefit of the coverage of WL\footnote{The other two are less affected because the subject of the rule (e.g., stop words) largely dictates which words it covers.}, its validity is generally lower due to under-representation of common words. 
    \item Although WL achieves highest coverage \textit{and} validity, it has $>500$ rules at $K{=}30$, with similar words having very different ranges, as shown in Fig.~\ref{fig:sst-baseline} -- a conglomerate (almost) impossible to make sense of. It also overfits, as the evaluation set validity is much lower than the construction set validity (which is 100\% by construction). 
    \item At $K{=}10$, only the stop word rule with random pick QF achieves validity $>80\%$, indicating that even the more careful practices are unreliable. 
\end{itemize}

\noindent All the drawbacks call for a principled way to derive robust model understanding with enforceable metric values (e.g. validity). As we demonstrate next, given a large construction set and automatic parameter tuning assistance, we can create such a \modelname{} rule union. Finally, as a meta-point, the above discussion above of various limitations would not be possible without the proposed \modelname{} formalization and metric definitions.

\paragraph{\modelname{} Construction}

We create a rule union consisting of nine rules, with target validity of 90\% and tune the sharpness accordingly.  Tab.~\ref{tab:sst-eoe-development} summarizes the individual and aggregate metrics. 

\begin{table}[!t]
    \centering
    \resizebox{\columnwidth}{!}{
    \begin{tabular}{rlrrr}\toprule
        Idx & Rule & Cov\% & Val\% & Shp\%\\\midrule
        1 & Negation & 1.2 & 89.5 & 65.1\\
        % 2 & Pos. adj & 3.2 & 90.9 & 40.4 & Rpl. 4\\
        2 & Strongly neg. adj & 3.2 & 91.6 & 83.5\\
        3 & Strongly pos. words & 5.1 & 91.9 & 40.0\\
        4 & Strongly neg. non-adj & 1.2 & 89.9 & 71.4\\
        5 & Person name & 2.4 & 90.9 & 28.4\\
        % 7 & Stop words & 47.5 & 91.6 & 18.2 & Rpl. 8-18\\
        6 & Stop words & 47.5 & 90.8 & 23.5\\
        7 & Zero-sentiment words & 17.1 & 90.0 & 15.6\\
        8 & Weakly pos. words & 15.4 & 91.2 & 11.3\\
        9 & Weakly neg. words & 5.7 & 91.7 & 31.4\\
        \midrule
        \multirow{2}{*}{\makecell{Un-\\ion}} & On construction set & 100 & 90.7 & 26.1\\
         & On evaluation set & 100 & 89.4 & 26.2\\
    \bottomrule
    \end{tabular}
    }
    \caption{Metrics for SST rules and rule union. }
    
    \label{tab:sst-eoe-development}
\end{table}

Clearly, high validity comes at the cost of low sharpness. Since (1 ${-}$ sharpness) is the probability that a random FEU has an explanation value within the behavior range, this around 90.7\% validity should be put into a context where the random baseline achieves a validity of around 75\%. In this sense, we attain only a crude understanding of the local explanations that misses many subtleties. 

Nonetheless, Rule 3 (strongly positive words) and Rule 6 (stop words) achieve better validity-sharpness trade-off than their counterparts created using the \textit{ad hoc} BG and QF methods above. Moreover, the WL rules cover all words seen in the analyzed instances -- analogous, in a sense, to our \modelname{} rule union. While the validity-sharpness trade-off is comparable between the two, ours has 100\% coverage due to the effectively ``catch-all'' Rule 7, while WL rules have less than 60\%. Most importantly, as our rule union is composed of nine semantically organized rules, it is much more interpretable than WL, which include more than 500 unpredictably varying rules (Fig.~\ref{fig:sst-baseline}). 

The fact that the \modelname{} rule union reveals the imprecision and limitations of our model understanding while still performing better than current practice emphasizes the need for more formal and quantitative model understanding, as well as the development of methods that are easier to \textit{understand}, in addition to being correct. Below, we highlight two sets of rules that quantitatively support or refute our intuition, and cover the rest in App.~\ref{app:sst-union}. 

\noindent \textit{\textbf{Rule 2, 3, 4, 8, 9: Sentiment-carrying words.}} We expect  a sentiment classifier to recognize sentiment-laden words. To test our intuition, we create rules for positive and negative words, and further split each set of words into two according to sentiment strength, resulting in four rules. For the two rules on strong words, we find that wide behavior ranges of [0.01, 1] and [-1, -0.01] are necessary to achieve 90\% validity, suggesting the looseness of the model understanding. However, we do observe that negative adjectives (but not positive ones) are modeled much better, where a range of [-1, -0.06] is sufficient for the same validity. Thus, we create a separate Rule 2, with very high sharpness of 84.2\%. For the two rules on words of weaker sentiment, even wider ranges of [-0.11, 1] and [-1, 0.05] are necessary. Since both ranges encroach upon the other side, the model often considers these words to have an impact opposite to their intrinsic meaning, but we fail to extract further understanding. In addition, negative rules are much sharper than positive ones, suggesting that the model considers a negative word to be stronger evidence for a negative prediction than its positive counterpart. 

\noindent \textit{\textbf{Rule 6: Stop words.}} While stop words (e.g., ``the'', ``of'') \textit{should} have negligible impact on prediction (and saliency values close to zero), a narrow behavior range of [-0.05, 0.05] only has 64\% validity. We create this rule for all stop words with 90\% target validity and use different ranges on different words for better sharpness. On average, we get [-0.07, 0.12], demonstrating that they can sometimes be more influential than even strong sentiment words. The ranges also tilt to the positive side, uncovering a grammaticality bias wherein prediction is more negative for grammatically incorrect sentences with stop words masked out by SHAP.

% \vspace{-0.5ex}
\subsection{Paraphrase Detection}
% \vspace{-0.5ex}
\label{sec:experiment-2}

\paragraph{Setup}

We use LIME explanations \citep{ribeiro2016should} for a fine-tuned BERT model \citep{devlin2018bert}, with 500 random test sentences as the construction set and the remaining $\approx$ 40k as the evaluation set. We remove the word sentiment feature but add the question ID (1 or 2) of each FEU.

\begin{table}[!t]
    \centering
    
    \resizebox{\columnwidth}{!}{
    \begin{tabular}{rlrrrl}\toprule
        Idx & Rule & Cov\% & Val\% & Shp\%\\\midrule
        1 & Matching words neg. pred & 11.7 & 90.9 & 39.5 \\
        2 & Matching words pos. pred & 12.4 & 90.3 & 38.6 \\
        3 & Non-matching words neg. pred & 18.7 & 90.0 & 35.5 \\
        4 & Question mark neg. pred & 5.2 & 90.2 & 36.5 \\
        5 & Question mark pos. pred & 3.8 & 90.0 & 23.1 \\
        6 & Stop words neg. pred & 22.3 & 90.0 & 32.8 \\
        7 & Stop words pos. pred & 12.6 & 90.5 & 12.5 \\
        8 & Negation words neg. pred & 0.3 & 90.0 & 36.0 \\
        9 & Negation words pos. pred & 0.1 & 95.7 & 7.2 \\
        10 & All else neg. pred & 4.0 & 92.1 & 23.5 \\
        11 & All else pos. pred & 8.8 & 90.3 & 5.7 \\
        \midrule
        \multirow{2}{*}{\makecell{Un-\\ion}} & On construction set & 100 & 90.3 & 29.3\\
        & On evaluation set & 100 & 90.0 & 29.1\\
        \midrule
        \multirow{2}{*}{\makecell{Word\\Avg}} & On construction set & 100 & 90.8 & 29.4 \\
         & On evaluation set & 82.3 & 84.4 & 29.4 \\
    \bottomrule
    \end{tabular}
    }
    
    \caption{Metrics for QQP rules and rule union. The last two rows are for the baseline at the end of Sec.~\ref{sec:experiment-2}. }
    
    \label{tab:qqp-eoe-development}
\end{table}

\paragraph{\modelname{} Construction}
QQP is a more complex domain than SST, since the label is the semantic equivalence of \textit{two} sentences. The metric values for the \modelname{} are summarized in Tab.~\ref{tab:qqp-eoe-development}. Below, we describe how expectations for the model are validated, but a hidden -- and somewhat surprising -- phenomenon is also uncovered. All other rules are documented in App.~\ref{app:qqp-union}. 

\noindent \textit{\textbf{Rule 1, 2: Matching words.}} Due to the nature of the task, we expect the model to rely heavily on matching words. For such a word $u$, defined as (proper) noun, verb, adjective or pronoun that has exactly one case-insensitive match $v$ in the other question, we expect similar saliency to their match due to symmetry, or formally its saliency $s_u\in [s_v-\alpha, s_v+\beta]$, where $\alpha$ and $\beta$ are lower and upper margins. This behavior function is non-constant, with output depending on the saliency values of other words in the sentence. 

For the same margin, FEUs for pairs of negative predictions have much higher validity than positive ones, so we split the rule into two based on the prediction. Despite a less than 1\% difference in sharpness (Tab.~\ref{tab:qqp-eoe-development}), we have $\alpha=\beta=0.07$ for the negative rule, but 0.18 for the positive rule, suggesting that the matching words make a much larger and more unpredictable contribution to positive predictions. Interestingly, all other rules had wider intervals for positive predictions as well.

\noindent \textit{\textbf{Rule 3: Non-matching words.}} Next we study model behaviors for non-matching words, defined analogously to matching ones. Following the previous split based on predicted label, we designed two rules. The negative rule has a reasonably sharp behavior range of [-0.35, 0.01] at 90\% validity. Given that LIME saliency is the linear regression coefficient on a neighborhood created by word erasure, we conclude that the \textit{presence} of these non-matching words mostly causes the prediction to tilt toward the non-paraphrase (i.e. negative) class, indeed a very reasonable behavior. However, we cannot find a range with 10\% sharpness at 90\% validity for the positive rule and thus discard it. 

With regard to the sharpness contrast by predicted label, one explanation is that the model defaults to a negative prediction, since many negative pairs consist of completely unrelated questions and the model decision is largely insensitive to input perturbations, leading to stable LIME coefficients. On the other hand, a positive prediction requires the cooperation of all parts of both questions. Depending on the exact sentence structure, the importance of each word to the match are different and hard to predict, which prevents the rules from being sharp. 

\paragraph{Word Average Baseline} Here, we introduce a new baseline as an ``automated'' version of WL rules in SST. Specifically, for each word in the construction set, we compute a behavior range around its average saliency, with sharpness of 29.4\% (matching that of our \modelname{} rule union). As Tab.~\ref{tab:qqp-eoe-development} shows, the resulting rule union is much worse than our manual one on both evaluation set coverage and validity, which is not surprising as the word saliency \textit{should} be more context-dependent, due to the matching mechanism of paraphrase detection. Moreover, with more than 2,000 constituent rules, the rule union barely qualifies as any sort of \textit{generalized} model understanding.

% \vspace{-0.5ex}
\section{Related Work}
% \vspace{-0.5ex}
\label{sec:related-work}

As discussed in Sec.~\ref{sec:intro}, explanation evaluation usually has a focus on correctness (or faithfulness) -- i.e., whether the explanation truly reflects the model's reasoning process. This includes sanity checks \citep{adebayo2018sanity}, proxy metrics \citep{samek2016evaluating, arras2019evaluating}, and explicit ground truth \citep{zhou2021feature}. 
The understandability issue has been much less studied, with the exception by \citet{zheng2021irrationality}, who proposed an evaluation specifically for rationale models \citep{lei2016rationalizing}. \modelname{}, however, addresses \textit{post hoc} explanations of general black-box models. 

In addition, a few prior works have attempted to capture the ``end-to-end'' utility of explanations: whether access to explanations leads to performance increase in certain tasks. \citet{hase2020evaluating} propose a model-teaching-human setup, subsequently extended by \citet{pruthi2020evaluating} into an automated evaluation procedure. \citet{bansal2021does} study whether explanations can improve human-machine teaming performance. While these studies report mostly negative results, pinpointing the root cause is difficult due to their end-to-end nature. Poor \textit{understanding} of the explanations may be a major reason, as indicated by \modelname{}.

Last, some authors have proposed methods for understanding model predictions beyond individual instances. For example, the anchor method \citep{ribeiro2018anchors} generates an explicit domain of applicability for each explanation, while \citet{lakkaraju2016interpretable} and \citet{lakkaraju2019faithful} proposed to learn ``patches'' of the input space specified by logical predicates. \modelname{} also emphasizes the need to understand models that generalizes across instances, and uses logical predicates in the formulation, but focuses on model understanding via \textit{explanations} instead of direct \textit{predictions}, which can capture a wider variety of \textit{behaviors} (e.g. the matching and non-matching behaviors of the QQP model). Furthermore, the fine-grained analysis of behaviors allows us to investigate whether models are ``correct for the correct reason.''

% \vspace{-0.5ex}
% \section{Beyond Feature Attributions}
% \vspace{-0.5ex}
% \label{sec:beyond}
% \input{8beyond}

\section{The Many Faces of Understandability}

The central thesis of this paper is quite simple and intuitive: in order to understand a model from local explanations, we need to understand those local explanations. While \modelname{} is the first framework to explicitly formalize and quantify the notion of understandability, we argue that it is connected to many often-discussed and desirable properties of explanation (further details in App.~\ref{app:alignment}). 

\paragraph{Human Alignment}
Users sometimes expect explanations that are aligned with their expectations. For example, the fact that highly salient words convey strong sentiment is taken as evidence for the quality of an explanation method by \citet{li2015visualizing}. In image classification, this concept is typically implemented as a pointing game between the high-saliency region and the segmentation mask of the predicted class \citep{fong2017interpretable}. However, alignment does not imply correctness, as the model could use any spurious correlations, which should be faithfully highlighted by the explainer. However, higher-alignment explanations are indeed \textit{more understandable}, since by definition they agree more with human intuition. Thus, an alignment-based evaluation can be considered as one of understandability. Nonetheless, understandability can also be achieved by correcting human expectations, e.g., users realizing that punctuations are actually important for predictions (contrary to expectations). 

\paragraph{Robustness}
It is often argued that explanations should be robust \citep{ghorbani2019interpretation} -- similar inputs should induce similar explanations. However, robustness can be at odds with correctness: if the model truly applies vastly different logic for two very close inputs -- such as a pair of inputs that only differ in the root feature of a decision tree -- then their explanations should be distinct, as they are routed down two different sub-trees. Nonetheless, slow-varying explanations are generally easier to understand than those that change erratically and unpredictably (independent of their correctness), and thus robustness is related to understandability. 

\paragraph{Counterfactual Similarity and Plausibility}
Counterfactual explanations \citep[e.g.][]{ross2021explaining} indicate how the input should change in order to alter the model prediction. Besides the success rate of achieving target prediction, they are often evaluated on similarity (the magnitude of input change) and plausibility (the naturalness of the changed input). Both properties can serve as proxies for understandability: it is easier to relate an input to another similar and natural input than to a totally different or abnormal one. However, App.~\ref{app:alignment} presents two cases where they should \textit{not} be similar or plausible but remain understandable, to highlight certain model behaviors.

% \vspace{-0.5ex}
\section{Discussion and Conclusion}
% \vspace{-0.5ex}
\label{sec:conclusion}

Traditionally, model explanations are evaluated on correctness (or faithfulness), i.e., whether they correspond to how models actually make predictions, e.g., reliance on spurious correlations \citep{zhou2021feature, adebayo2021post}. Such evaluation, however, does not answer the equally important question of whether these (presumably correct) explanations are understandable.
Even faithful explanations can lead users into error, if misunderstood (e.g., trusting a model incorrectly).

In a sense, the most correct explanation for an input is the literal trace of model computation, but it is also arguably the least understandable (or useful). As we abstract away from low-level details and use higher-level concepts such as word sentiment, the resulting explanation loses correctness but gains understandability. At the other extreme are explanations that are trivially understandable but completely wrong, such all attribution values being 0 (i.e., no feature impacts the model prediction). Thus, a trade-off often occurs between these two desiderata, and we need to choose a sweet spot. 

Concretely, we propose \modelname{} rules and rule unions, along with three quality metrics to formalize and evaluate understandability. Such rigorous investigations stand in contrast to current \emph{ad hoc} practices, which are prone to yielding unreliable and coarse model understanding. For SST and QQP datasets, \modelname{} demonstrates that our model understanding is quite limited and imprecise, even with very reasonable explanations. \emph{Being aware of this is an asset}. While \modelname{} helps us to recognize that our understanding is incomplete, it still helps uncover unexpected model behaviors that warrant further investigation.

\section*{Limitations and Ethical Impacts}
\subsection*{Limitations}

One notable requirement of \modelname{} is the extensive human involvement in constructing and optimizing its rules. However, this process is necessary, as the alternative of generalizing from a few explanations has various flaws, depicted in Fig.~\ref{fig:1} and Fig.~\ref{fig:current-practice-sst}. Practically, we spent about 3 hours on each rule union in Sec.~\ref{sec:experiment}, and our effort was streamlined by the systematic process and GUI presented in Sec.~\ref{sec:gui}, which could be further improved by methods that automatically propose candidate rules. 

In addition, another area requiring human involvement is the FEU feature definitions, which are often domain-dependent: both the sentiment score and the matching word features reflect the nature of the tasks. Other features may be necessary for other tasks. For example, in question-answering, one important FEU feature could be the kind of interrogative word used in the question (e.g., ``what'' vs. ``when'' vs. non-interrogative words). If important features are missed, the quality of the \modelname{} rules -- and, hence, the model understandings -- will suffer accordingly. 

Last, the difficulty of obtaining overall high-quality model understanding may result from the fundamental limitations of word-level attribution-based explanations, which cannot account for higher-level interactions. \modelname{} could aid in the development of new explanation methods that are easier for humans to understand. As a first step, we explore defining and evaluating model understanding obtained from instance-based explanations with whole input as FEUs. App.~\ref{app:ibe} details the investigation, which raises questions such as the reliability of such explanations. 

\subsection*{Ethical Impacts}

As interpretability methods are increasingly deployed for quality assurance, auditing and knowledge discovery purposes, it is important to ensure the legitimacy of any conclusions drawn from explanations. While the correctness of these explanations is often studied, we argue their understandability should be equally emphasized, and evaluations with our newly proposed \modelname{} framework and GUI reveal many problems of existing \textit{ad hoc} procedures. Thus, a more careful treatment on the understandability aspect is necessary for well-calibrated model understandings and responsible model deployment in the real world.

\section*{Acknowledgment}
This research is supported by the National Science Foundation (NSF) under the grant IIS-1830282. We thank the reviewers for their efforts. 

\bibliographystyle{acl_natbib}
\bibliography{references}

% \newpage
\onecolumn
\appendix

\section{Real World Use Cases for Explanations}
\label{app:real-world}
Here, we discuss several scenarios in which people use local explanations to understand models, and argue that people invariably derive \textit{generalized} model understanding from these explanations. 

\subsection{Spurious Correlation Identification}
\label{app:real-world-spurious}
Natural datasets can contain many spurious correlations. For example, in a COVID-19 chest X-ray dataset, most positive images (i.e., patients diagnosed with COVID-19) come from a pneumonia-specializing hospital and contain a watermark of the hospital name, while most negative images from other hospitals do not. Thus, a model could achieve very high accuracy by simply detecting the watermark rather than genuine medical signals. Similar spurious correlations could also be present in the text domain, such as the correlation between an exclamation mark and the positive sentiment class, or between the word ``\textit{not}'' and the contradiction class in natural language inference. 

It is crucial for people to be aware of the shortcuts that models may take, and one possible way to highlight such behaviors is via feature attribution, which in the examples above would assign an abnormally high score to the watermark region, exclamation mark, or the word ``\textit{not}.'' Assuming the explanations do indeed exhibit such patterns, when people claim a model relies on spurious correlation, they mean this in a general sense: for example, the model is likely to focus on the watermark in \textit{any} image that contains it, rather than in only a specific set of images.

\subsection{Fairness Assurance}
Similar to spurious correlation features, other features should not have a high impact, but for reasons of fairness. For example, decisions made by a loan approval model should not be affected by gender\footnote{There could be other features that correlate with gender, such as job title, but we ignore such possibilities for simplicity.}, therefore the gender feature should not have a high attribution score. 

If we observe that the gender of one applicant heavily impacts the model’s decision, we may suspect the model is discriminative; conversely, observing that it has minimal impact could increase our assurance of the model's fairness. However, such single-instance observations are fundamentally exploratory, and claims about the model’s fairness or discrimination must be established using a \textit{population} of instances to determine whether the trend persists generally.

\subsection{Model-Guided Human Learning}
In some cases, a very accurate and ``super-human'' model could be a source for knowledge discovery. Consider the task of early-stage cancer detection from CT scans, which is challenging for doctors. If a label is generated from follow-up visits tracking whether patients develop cancer after a certain number of years, a model achieving better test accuracy than doctors  is likely to use certain cues that would be missed by humans or not known to be linked to cancer. 

For these models, explanation methods such as saliency maps could be used to help doctors make better diagnoses, or assist scientists in the creation of new pathological theories. Similarly to the above two use cases, \textit{generalized} model understanding  across different inputs are necessary, because doctors need to apply what they have learned to new patients, and scientists require new theories to hold broadly.

% \section{Intersection Mode Composition}
% \label{app:intersection-mode}
% \input{appendix/1intersection}

% \section{Graphical User Interface (GUI) for \modelname{} Construction}
% \label{app:gui}
% \input{appendix/2gui}

\section{Additional Evaluation Details}
\label{app:experiment}

Tab.~\ref{tab:exp-overview} summarizes the key parameters of our experiment. Both saved models are publicly accessible from Huggingface Hub, and the model names in the table are links to the respective model checkpoints. For normalization, we divided all explanation values for all test set instances by a single scaling factor such that the maximum magnitude of new explanations is 1. 

\begin{table}[!htb]
    \centering
    \begin{tabular}{llllll}\toprule
        Task & Dataset & Model & Acc. & F1 & Explainer\\\midrule
        Sentiment & SST-2 & \href{https://huggingface.co/textattack/roberta-base-SST-2}{RoBERTa} & 95.6\% & 0.957 & SHAP\\
        Paraphrase & QQP & \href{https://huggingface.co/zyl1024/bert-base-cased-finetuned-qqp}{BERT} & 90.7\% & 0.875 & LIME\\\bottomrule
    \end{tabular}
    \caption{A summary of tasks, models (fine-tuned on respective datasets), and explainers for the two case studies. }
    \label{tab:exp-overview}
\end{table}

\subsection{SST Sentiment Classification}
\label{app:sst-experiment}

\begin{figure}[!htb]
    \centering
    \includegraphics[width=\textwidth]{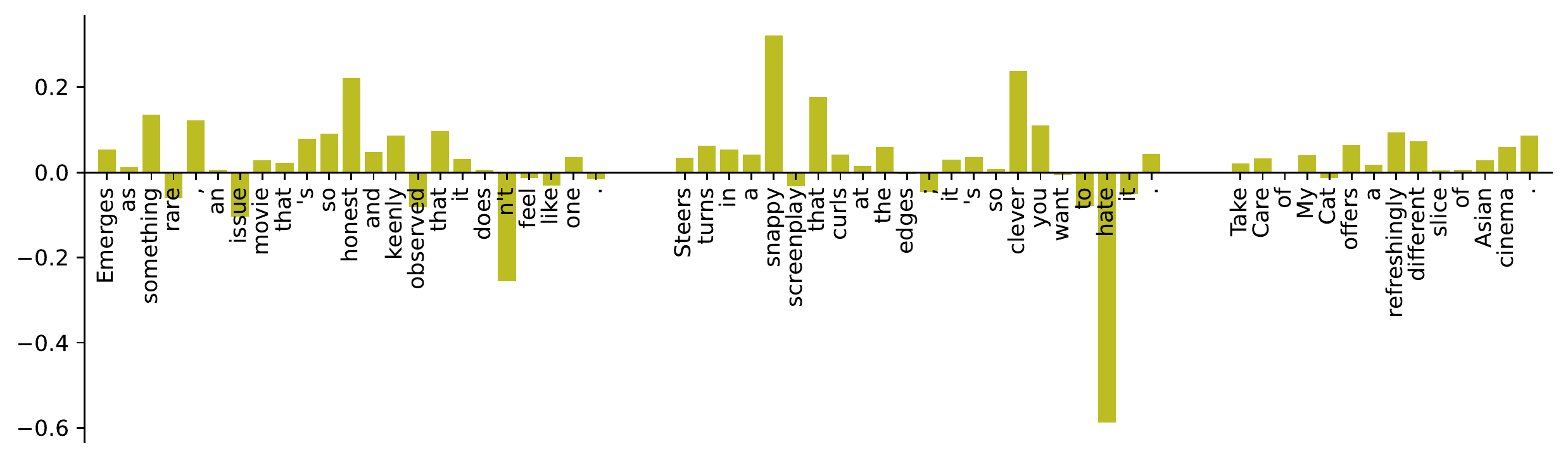}
    \caption{SHAP explanation visualization for three SST inputs. }
    \label{fig:sst-shap-vis}
\end{figure}

For the explainer, we used the PartitionSHAP algorithm implemented by the \Verb[fontfamily=cmtt]+shap+ repository\footnote{\url{https://shap.readthedocs.io/en/latest/}}. Fig.~\ref{fig:sst-shap-vis} shows the explanations on three sentences (after normalization). 

\subsubsection{Details of Current Practices}
\label{app:bg-qf-wl}

Here, we provide an extended description of the three current practices, and how they are applied on the handful of selected examples, collectively called the ``sample'' below. 

The first method, ``belief guided'' (BG), represents the practice wherein the user has some expectations (or beliefs) about the attributions of certain words, and modifies (or updates) them after observing explanations on  some actual test inputs. It operates differently for the two rules on  positive-sentiment and stop words, as follows. 
\begin{enumerate}[leftmargin=*]
    \item For positive-sentiment words, the prior belief is that a word with a higher sentiment score (one of the FEU features provided by the SST dataset) should also receive more positive attribution. This leads to a rule that applies to all words with a sentiment score greater than $\alpha$, and has a behavior function that outputs a constant range of $[\beta, 1]$ (recall that SHAP attribution values are normalized to $[-1, 1]$ range). It  then computes the value of $\alpha$ as the mean sentiment score and $\beta$ as the mean attribution value for all words in the sample with positive sentiment scores. 
    \item For the stop words -- defined as those with parts of speech AUX, DET, ADP, CCONJ, SCONJ, PRON, PART, and PUNCT -- it  has a prior belief that they should have a attribution value range of [-0.05, 0.05] (i.e., not important to model prediction), and computes the observed attribution range $[\alpha, \beta]$ for stop words in the sample. The final behavior range as predicted by the behavior function of this rule is the average of these two: $[-(0.05+\alpha)/2, (0.05+\beta)/2]$. 
\end{enumerate}

The second method, ``quantile fitting'' (QF), represents the practice wherein the user fully follows the observed data without any prior beliefs. Specifically, for a set of words, it  collects all attribution values for words within the set and then creates a rule that applies to this set, with the behavior function predicting a constant range of 5\% to 95\% quantile of these attribution values. For the two rules for positive-sentiment and stop words, the set of words (and hence the applicability functions) is  defined in the same way as for the BG method above. 

The last method, ``word-level'' (WL), can be considered a more extreme version of QF, where the user not only lacks any prior expectations for the explanations but also considers each word individually. For example, if the user observes that the word ``brilliant'' has an attribution value  of 0.5 in one sentence and the word ``fantastic'' has attribution of 0.8 in another, they would \textit{not} conclude that other,  similarly positive words would have attributions approximately  within the range of $[0.5, 0.8]$. Specifically, for every distinct word $w$ in the sample, this method builds a rule that applies only to that word, with a constant behavior function that outputs a range of $[\min(s_w) - 0.03, \max(s_w) + 0.03]$, where $s_w$ is the list of attributions received by different occurrences of $w$. In many cases, especially given a small sample, word $w$ only appears once, in which case $s_w$ is a list containing only that attribution value.

Tab.~\ref{tab:current-practice-sst} presents the metric values of the above methods. Fig.~\ref{fig:current-practice-sst} of Sec.~\ref{sec:experiment-1} depicts a graphic summary.

\begin{table}[!htb]
    \centering
    % \vspace{-0.15in}
    % \resizebox{0.65\textwidth}{!}{
    \begin{tabular}{rc cc cc c}\toprule
          & & \multicolumn{2}{c}{belief-guided} & \multicolumn{2}{c}{quantile-fitting} & word-level\\
      $K$ & pick & positive & stop word & positive & stop word & seen words \\\cmidrule(lr){1-2} \cmidrule(lr){3-4} \cmidrule(lr){5-6} \cmidrule(lr){7-7}
        \multirow{3}{*}{1}  & SP & 10, 72, 50 & 49, 45, 65 & 10, 63, 44 & 49, 63, 45 & 28, 51, 61 \\
        & RND $\mu$ & 12, 63, 57 & 49, 58, 53 & 12, 45, 56 & 49, 73, 33 & 17, 41, 68\\
        & RND $\sigma$ & \x{}6, 25, 27 & \x{}0, \x{}9, \x{}6 & \x{}6, 32, 29 & \x{}0, 24, 21 & \x{}6, 12, 10\\
        \cmidrule(lr){1-2} \cmidrule(lr){3-4} \cmidrule(lr){5-6} \cmidrule(lr){7-7}
        \multirow{3}{*}{10} & SP & 10, 61, 61 & 49, 47, 63 & 10, 78, 34 & 49, 72, 38 & 49, 66, 48\\
        & RND $\mu$ & 10, 71, 52 & 49, 56, 56 & 10, 75, 32 & 49, 84, 25 & 41, 68, 48\\
        & RND $\sigma$ & \x{}0, \x{}6, \x{}7 & \x{}0, \x{}4, \x{}4 & \x{}0, \x{}9, 10 & \x{}0, \x{}3, \x{}2 & \x{}2, \x{}1, \x{}2\\\cmidrule(lr){1-2} \cmidrule(lr){3-4} \cmidrule(lr){5-6} \cmidrule(lr){7-7}
        \multirow{3}{*}{30}  & SP & 10, 64, 59 & 49, 50, 60 & 10, 88, 17 & 49, 82, 29 & 57, 73, 42\\ 
        & RND $\mu$ & 10, 66, 56 & 49, 57, 55 & 10, 82, 26 & 49, 86, 24 & 51, 78, 39\\
        & RND $\sigma$ & \x{}0, \x{}4, \x{}5 & \x{}0, \x{}1, \x{}2 & \x{}0, \x{}6, \x{}7 & \x{}0, \x{}2, \x{}2 & \x{}2, \x{}3, \x{}2
        \\\bottomrule
    \end{tabular}
    % }
    % \vspace{-0.1in}
    \caption{Coverage, validity, and sharpness (percentage) of model understanding with \textit{ad hoc} current practice. ``SP'' refers to the submodular pick procedure, and ``RND'' refers to the random sampling procedure. The latter also shows mean $\mu$ and stdev $\sigma$ across five runs. }
    % \vspace{-0.25in}
    \label{tab:current-practice-sst}
\end{table}

\subsubsection{Complete Rule Union Description}
\label{app:sst-union}

\noindent Below, we present the details of the construction process for rules not discussed in Sec.~\ref{sec:experiment-1}. 

\begin{itemize}[leftmargin=*]
\item \textit{\textbf{Rule 1: Negation words have negative saliency.}} We found that negation words -- \textit{not}, \textit{n't}, \textit{no}, \textit{nothing} and those with NEG dependency tag -- almost invariably receive (sometimes highly) negative saliency, regardless of the sentence label or sentiment of the word being modified. We create a rule that predicts a constant behavior range $[-1.0, 0.002]$, with 89.5\% validity and 65.1\% sharpness. Although the validity is under our 90\% target, we found that to make it higher, the upper limit of the behavior range needs to be 0.1, which results in an extremely low sharpness of 11\%. Thus, we decided against it. 

\item \textit{\textbf{Rule 5: Person names have positive saliency.}} During our initial inspection, we found several cases where the name of a person (e.g. director or actor) have positive saliency values. Thus, we create this rule from the NER tag, covering 2.3\% of words. However, after parameter tuning, we found that while many of the words have positive saliency, the correct characterization is that they all have small saliency values, as a behavior range of $[-0.06, 0.1]$ achieves 91.6\% validity. However, since SHAP saliencies are mostly concentrated around 0, this range achieves a meager sharpness of 26.8\%. Despite this, we still decide to keep it. 

\item \textit{\textbf{Rule 7: Zero-sentiment words have small saliency.}} Besides stop words, we should expect words that do not carry sentiment, such as most nouns and verbs (e.g., \textit{movie} and \textit{get}), to have small saliency magnitudes. Due to the wide range of words applicable under this rule, we choose the saliency range to be $[-0.15, 0.15]$ for $\geq 90\%$ validity, but this range yields lowest sharpness of 13.5\%. 

\item \textit{\textbf{Rule 8, 9: Weakly positive/negative words have weakly positive/negative saliency.}} Finally, we set up two rules to capture words that have sentiment of neither zero (covered by Rule 19) nor high-magnitude (covered by Rule 3 -- 5). To achieve 90\% validity, we require a behavior range of $[-0.11, 1]$ for weakly positive words and $[-1, 0.05]$ for weakly negative words, unfortunately again with quite low sharpness. Notably, both ranges need to ``spill over'' to the other side of zero for the required validity. 
\end{itemize}

\subsection{QQP Paraphrase Detection}
\label{app:qqp-experiment}

\begin{figure}[!htb]
    \centering
    \includegraphics[width=\textwidth]{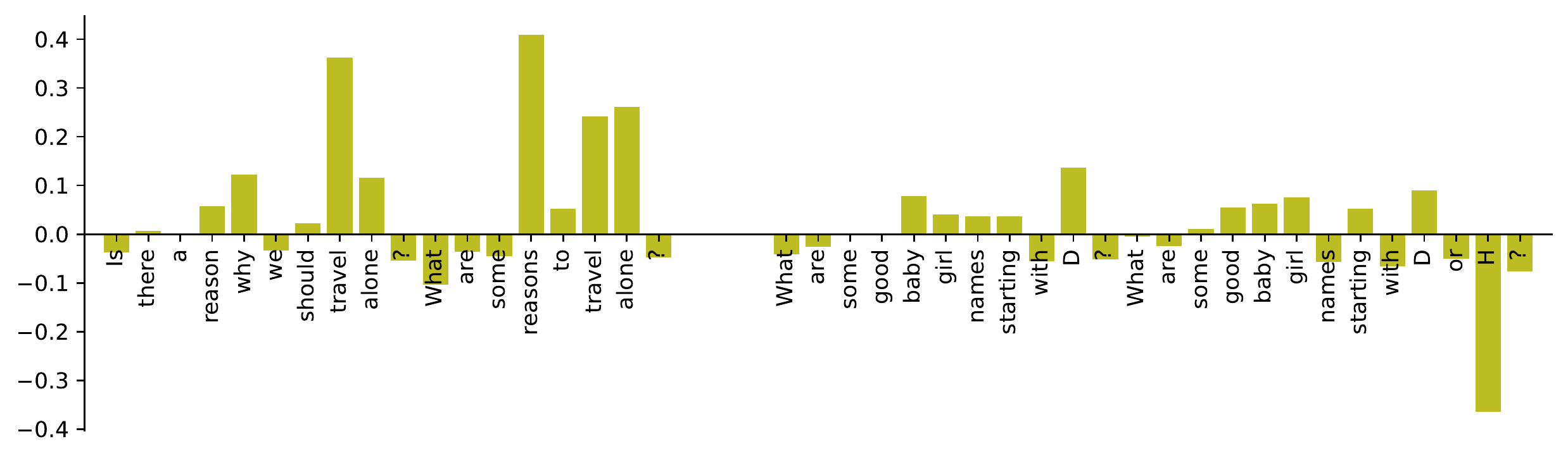}
    \caption{LIME explanation visualization for two QQP pairs. }
    \label{fig:qqp-lime-vis}
\end{figure}

For the explainer, we used the LIME algorithm implemented by the \Verb[fontfamily=cmtt]+lime+ repository\footnote{\url{https://github.com/marcotcr/lime/}}. Fig.~\ref{fig:qqp-lime-vis} depicts the explanations on two pairs (after normalization). 

\subsubsection{Complete Rule Union Description}
\label{app:qqp-union}

\noindent Below, we present details of the construction process for rules not discussed in Sec.~\ref{sec:experiment-2}. 

\begin{itemize}[leftmargin=*]
    \item \textit{\textbf{Rule 4, 5: Saliency for trailing question marks.}} Since the dataset is composed of pairs of questions, the vast majority of sentences conclude with question marks. These should be purely decorative and syntactic, and so should have small saliency, similar to stop words. However, we observe that the saliencies assigned to them for positive and negative predictions are very different, so we create two rules for these two cases. With a 90\% validity target, the saliency range is [-0.04, 0.03] for negative predictions and [-0.07, 0.06] for positive predictions. Again, the saliencies for positive predictions demonstrate more variation than those for negative ones. 
    \item \textit{\textbf{Rule 6, 7: Saliency for stop words.}} Similar to SST, we use these two rules to ensure stop words should \textit{not} be influential. We split the stop word group into finer segments by part of speech, to achieve higher sharpness. On average, the range is [-0.07, 0.03] for negative predictions and [-0.09, 0.1]for positive predictions, which again demonstrate a much higher degree of variation. 
    \item \textit{\textbf{Rule 8. 9: Saliency for negation words.}} In the SST case, we found that negation words typically have negative saliency regardless of the sentiment label, and test whether this holds for QQP as well. Following on our previous findings, we use two rules to separately model inputs of positive and negative predictions. We find that the range is [-0.1, 0.24] for positive predictions and [-0.21, 0.01] for that for negative predictions. Curiously, the same negative saliency trend is preserved here as well, but only for inputs with negative predictions. 
    \item \textit{\textbf{Rule 10, 11: Saliency for everything else.}} Finally, we designed two lowest-precedence ``catch-all'' rules to complete the coverage. The range for positive prediction FEUs is [-0.13, 0.25]. For negative prediction inputs, we find that breaking them according to different parts of speech (nouns, verbs, adjectives, and everything else) is helpful, with verbs having a particularly narrow saliency range of [-0.05, 0.05]. On average, the saliency range is approximately [-0.09, 0.05]. 
\end{itemize}

\section{Understandability as a Unified Theme}
\label{app:alignment}

In this section, we elaborate on how understandability is the unified theme behind many properties of explanations that seem ``orthogonal'' to correctness. Specifically, we discuss three properties: human alignment , robustness, and counterfactual similarity and plausibility. 

\subsection{Human Alignment}
\begin{figure}[!htb]
    \centering
    \includegraphics[width=0.4\textwidth]{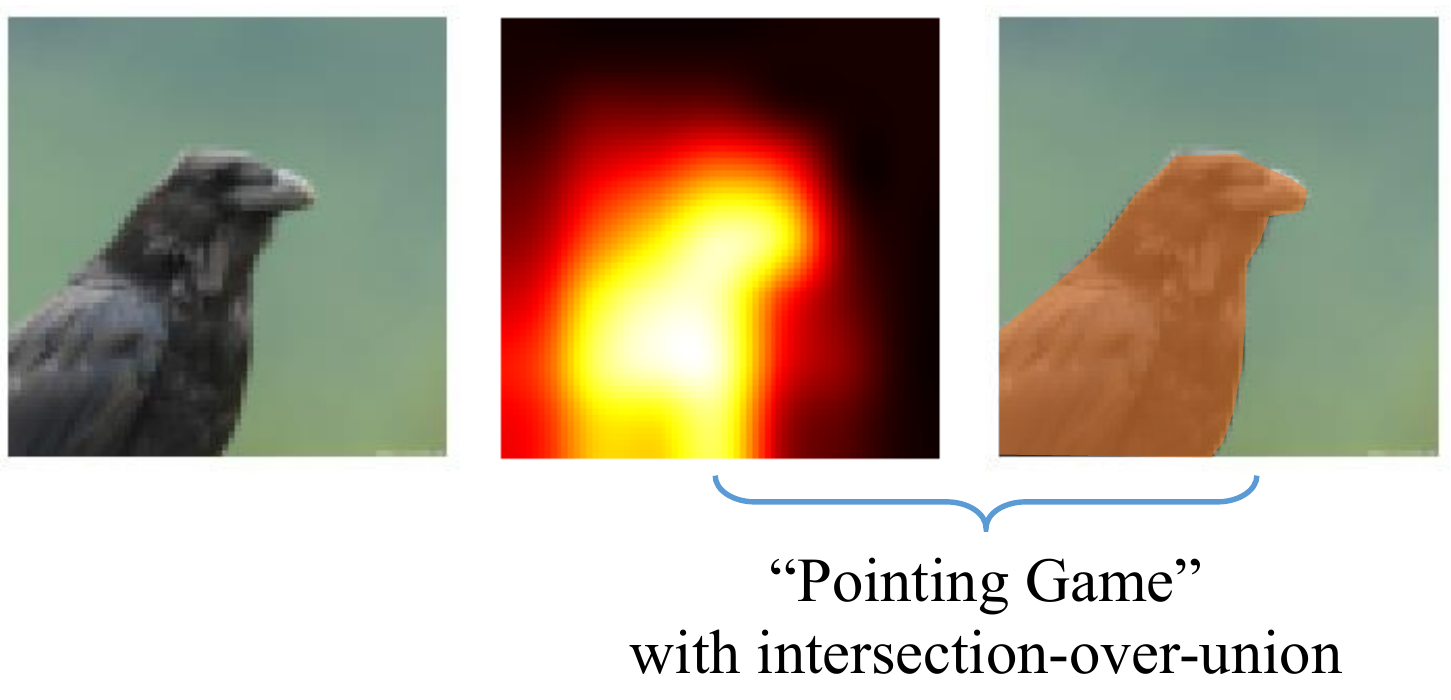}
    \caption{A pointing game used to quantify human alignment for visual explanations.}
    \label{fig:iou}
\end{figure}

Many prior works have assessed how much explanations agree with human expectation. For example, \citet{li2015visualizing} observed that the word ``hate '' contributes the most to a negative sentiment prediction in many inputs, and used it  to argue the explanation  is correct. In a similar sentiment classification task, \citet{bastings2019interpretable} used the high degree of overlap between the extracted rationale and strong-sentiment words to argue the superior quality of a neural rationale model \citep{lei2016rationalizing}. In computer vision, this alignment is often implemented as a pointing game that computes the intersection-over-union (IoU) metric between the salient region and the semantic segmentation mask of the predicted class \citep{simonyan2013deep, fong2017interpretable}, as shown in Fig.~\ref{fig:iou}. For a model that predicts breast cancer onset using patients’  genetic information, \citet{covert2020understanding} demonstrated that many of the influential genes identified by their explainer were indeed known to be associated with the disease.  

As discussed in App.~\ref{app:real-world-spurious}, models could use any unexpected spurious correlation, such as the green background in Fig.~\ref{fig:iou}. For these models, correct explanations should have low alignment scores. When correctness (or faithfulness) is the sole desideratum of interpretability methods, it is unclear what purposes these alignment evaluations serve. Some authors \citep[e.g.][]{jacovi2020towards} have even argued they are fundamentally misleading and flawed in nature as they focus on \textit{plausibility}, which is sometimes at odds with the goal of correctness. 

However, from the perspective of \textit{understandability}, high-alignment explanations are arguably very understandable, simply because they align closely with human expectation. Thus, given the same level of correctness, a higher-alignment explainer may be preferable.

\subsection{Robustness}
Besides human alignment, robustness -- i.e. that similar inputs should have similar explanations -- is also argued to be a favorable property for explanation. For example, \citet{ghorbani2019interpretation} argued that explanations are fragile due to their adversarial vulnerability, \citet{alvarez2018robustness} empirically estimated the Lipschitz constant for many explainers, and \citet{alvarez2018towards} proposed an inherently interpretable model that is explicitly regularized for explanation robustness.

\begin{figure}[!htb]
    \centering
    \includegraphics[width=0.25\textwidth]{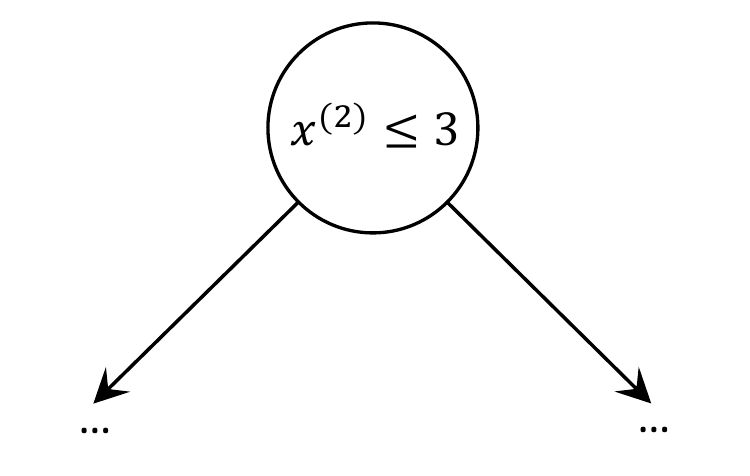}
    \caption{A decision tree that splits on the second feature at the root node. }
    \label{fig:dt}
\end{figure}

Robustness generally conflicts with correctness. If, for two inputs, the model is using distinct reasoning patterns, the correct explainer should faithfully report distinct explanations for them. One straightforward example is the decision tree model shown in Fig.~\ref{fig:dt}, where the root node splits on the second feature at a threshold value of 3. For two inputs $x_1$ and $x_2$ that agree on all features except the second one, with $x_1^{(2)}=2.99$ and $x_2^{(2)}=3.01$, since they are sent down two different sub-trees at the very beginning, the model is likely to use for totally different features. 

Nonetheless, as implicitly argued by the works above, erratic model behaviors are less understandable because they make it more difficult to identify generalizable patterns compared with slowly varying explanations in the input space. Thus, robustness is another aspect of the same understandability desideratum. 

\subsection{Counterfactual Similarity and Plausibility}
Unlike feature attribution explainers that assign importance to individual features, counterfactual (CF) explainers \citep[e.g.,][]{ross2021explaining} directly generate whole inputs but for a target predicted class. Thus, a CF explanation indicates how to cross the decision boundary from the input. 

Naturally, the fundamental requirement of CF explanations is achieving the target prediction, which is  typically known as validity. However, this is trivially satisfiable by simply finding a training instance with the target prediction, along with other ways such as creating adversarially perturbed or nonsensical inputs. Thus, two additional requirements are often enforced: similarity and plausibility. The former says  that the CF explanation should be close to the original input (with regard to, for example, edit distance), and the latter says the CF explanation should be plausible, or natural. Tab.~\ref{tab:cf} depicts various CF explanations and their satisfaction of the three requirements. 
\begin{table}[!htb]
    \centering
    \begin{tabular}{ccccc}\toprule
    \multicolumn{5}{l}{Input: This restaurant is the best I have been, with especially great food.}\\\midrule
        CF & Type & Val. & Sim. & Plau. \\\midrule
        \makecell[l]{This restaurant is the \textit{worst} I have\\been, with especially \textit{terrible} food.} & ``good'' CF & \cmark & \cmark & \cmark\\[0.5ex]\hdashline\noalign{\vskip 0.5ex}
        \makecell[l]{Rude service!} & \makecell{training set\\look-up} & \cmark & \xmark & \cmark\\[0.5ex]\hdashline\noalign{\vskip 0.5ex}
        \makecell[l]{This \textit{resturant} is the best I have\\been, with especially great food.} & \makecell{adversarial\\typo injection} & \cmark & \cmark & \xmark\\[0.5ex]\hdashline\noalign{\vskip 0.5ex}
        \makecell[l]{Fjwpeaf fawekl fka erj sfdlk erjlm\\adl erio fd} & \makecell{nonsensical\\inputs} & \cmark & \xmark & \xmark\\\bottomrule
    \end{tabular}
    \caption{CF explanations that are all valid but differ in similarity and plausibility metrics. }
    \label{tab:cf}
\end{table}

Validity for CF can be considered as the correctness analogy for feature attribution, but the purposes of similarity and plausibility are not readily apparent. As CF explanations represent ways to cross the decision boundary, people need to meaningfully understand how the CF instance is related to the original input. It is difficult to relate two dissimilar instances, and an implausible CF instance is generally unexpected. Thus, similarity and plausibility are required to make CF explanations more understandable.  

Interestingly, if our true goal is the understandability of the relationship between the input and its CF explanation, there are cases where similarity or plausibility is \textit{not} desirable. First, consider a sentence length classifier that predicts positive for sentences of at least 10 words, and negative otherwise. Given an input of three words, the CF explainer should generate \textit{dissimilar} CF instances of at least 10 words in order to correctly illustrate the decision boundary, while instances of even more words would be helpful for understanding the ``at least 10 words'' logic. Second, consider a classifier trained on a typo-free dataset and having high probability of making mistakes on inputs that contain typos. To illustrate this behavior, CF explanations should contain randomly (not adversarially) injected typos, which are \textit{implausible}, but useful as long as the typo injection is understood by people.

\section{Additional Details on Instance-Based Explanations}
\label{app:ibe}

In this section, we describe our initial attempt at extending the \modelname{} framework to another type of explanations: instance-based explanations (IBE). The IBE for an input $x$ is a set of instances and their predictions $\{(x_i, \widehat y_i)\}$, where $x$ and $x_i$ are semantically related (e.g., negation). We define $\widehat y_i$ as the predicted probability of positive class. 

We use \textsc{Polyjuice} \citep[PJ,][]{wu2021polyjuice} to generate instances of three semantic operations. \textit{Entity change} replaces a proper noun (e.g. actor name) with another using ``lexical'' mode of PJ. \textit{Minor insert} adds a short text to the sentence using ``insert'' mode. \textit{Negation} generated a negated version of the input using ``negation'' mode. 
For each operation type, our expectation for model behavior is formalized as a range $b(\widehat y)$ on $\widehat y$. We expect the prediction to be unchanged by the first two operations allowing for a margin of 0.05, but changed to the other side of 0.5 by negation. We then define validity $\nu=\mathbb E_{\widehat Y, \widehat Y_i} \left[\mathbbm 1_{\widehat Y_i\in b(\widehat Y)}\right]$ and sharpness $\sigma = 1 - \mathbb P_{\widehat Y} [b(\widehat Y)]$ analogously. 

\begin{table}[!t]
    \centering
    \begin{tabular}{lrcc}\toprule
        Type & $b(\widehat y)$ & $\nu$ & $\sigma$\\\midrule
        Entity change & $\widehat y \pm 0.05$ & 91.4 & 56.5 \\
        Minor insert & $\widehat y \pm 0.05$ & 89.1 & 57.3 \\
        Negation & $\mathrm{other\text{-}side}(\hat y)$ & 30.4 & 50.0  \\[0.5ex]\hdashline\noalign{\vskip 0.5ex}
        Negation & $\mathrm{same\text{-}side}(\hat y)$ & 69.6 & 49.9 \\
        Negation ($\leq$ 6 words) & $\mathrm{other\text{-}side}(\hat y)$ & 56.2 & 49.7 \\ \bottomrule
    \end{tabular}
    \caption{Instance-based explanation metrics on SST. }
    
    \label{tab:ibe}
\end{table}

\begin{table}[!b]
    \centering
    \begin{tabular}{p{7.5cm}p{7.5cm}}\toprule
        Input sentence & ``Negated'' sentence \\\midrule
         Human Nature initially succeeds by allowing itself to go crazy , but ultimately fails by spinning out of control . & Human Nature initially succeeds by allowing itself to go crazy , but ultimately fails by not coming to consciousness . \\[0.5ex]\hdashline\noalign{\vskip 0.5ex}
         This may be the dumbest , sketchiest movie on record about an aspiring writer 's coming-of-age . & This may be the dumbest , sketchiest movie on record , not an aspiring writer 's coming-of-age .\\[0.5ex]\hdashline\noalign{\vskip 0.5ex}
         Before long , the film starts playing like General Hospital crossed with a Saturday Night Live spoof of Dog Day Afternoon . & Before long , the film starts playing like nothing crossed with a Saturday Night Live spoof of Dog Day Afternoon .\\\midrule
         A startling and fresh examination of how the bike still remains an ambiguous icon in Chinese society . & A startling and fresh examination of how the bike still seems to be an ambiguous icon in Chinese society .\\[0.5ex]\hdashline\noalign{\vskip 0.5ex}
         Never engaging , utterly predictable and completely void of anything remotely interesting or suspenseful . & Not engaging , utterly predictable and completely void of anything remotely interesting or suspenseful .\\[0.5ex]\hdashline\noalign{\vskip 0.5ex}
         Between the drama of Cube ? & Are there no interesting problems? \\[0.5ex]\hdashline\noalign{\vskip 0.5ex}
         Tailored to entertain ! & No tails !\\\bottomrule
    \end{tabular}
    \caption{Failure cases of \textsc{Polyjuice} negations. The first half shows examples where the negation is irrelevant to the sentiment. The second half includes examples where the negation fails to appear. }
    \label{tab:ibe-negation}
\end{table}

Tab.~\ref{tab:ibe} summarizes the results. While our expectation is mostly confirmed for entity change and minor insert, it is notably violated in the case of negation, with only 30.4\% validity, indicating model prediction is on the \textit{same} side 69.6\% of time. Upon further evaluation, we find that validity drops with sentence length, with short sentences of six words or fewer having much higher validity (for other-side). Since the PJ rewriting model is learned rather than manually defined and negation is more complex than the other two operations, there are two failure modes, as presented in Tab.~\ref{tab:ibe-negation}. In the first, a negation is applied to the input sentence, but on a part irrelevant to the sentiment. In the second, the generated sentence is not a negation of the input by any reasonable standard. 

These examples highlight the importance of clearly defining the operation: rather than a generic negation, we would need the negation to happen on the ``sentiment-carrying'' part. It is also crucial to ensure that the generator is of a high quality in order to minimize the chance of generating nonsensical outputs. Despite many advances in generative language modeling, it have been shown to be undesirable in many ways \citep[e.g.,][]{holtzman2019curious}, all of which affect the quality of the counterfactual explanation.

At a high level, IBE explains the local prediction by illustrating ways to cross (e.g., negation) or not cross (e.g., entity change) the decision boundary in the (very) high-dimensional input space. However, as the negation case indicates, we must be careful about the exact definition of the rewriting (e.g., negating any part of the input or the ``sentiment-carrying'' part only), as it could have a significant impact on the conclusion. Furthermore, it is difficult for any rewriting mechanism to achieve 100\% validity due to the high dimensionality, the multitude of possible ways of rewriting, and the imperfection of the model. Focusing only on the mistakes (or ignoring them altogether) yields incomplete model understanding. Instead, the validity metric, which indicates the \textit{generalized} model behavior, should be used to. 

There are many potentially fruitful directions for future work. First, the quality of instances obviously depend on the generative models, which, while impressive, are known to be flawed in many ways \citep[e.g.,][]{holtzman2019curious, nadeem2020stereoset, wolfe2021low}. Second, each rule essentially covers the entire input space. Partitioning the input space in some way may allow for identification of both more and less consistent areas, which is makes the applicability function much more difficult to define as it now takes whole sentences rather than individual words. Finally, unlike feature attribution, which conveys the single notion of ``importance,'' different instances of the same input can reveal different aspects of model behavior, calling for a potentially different definition of coverage, which measures completeness of understanding.

\end{document}